\documentclass[lettersize,journal]{IEEEtran}
\usepackage{amsmath,amsfonts}
\usepackage{algorithmic}
\usepackage{algorithm}
\usepackage{multirow} 
\usepackage{array}
\usepackage[caption=false,font=normalsize,labelfont=sf,textfont=sf]{subfig}
\usepackage{textcomp}
\usepackage{stfloats}
\usepackage{url}
\usepackage{verbatim}
\usepackage{graphicx}
\usepackage{cite}
\usepackage{booktabs}
\usepackage{balance}
\begin{document}

\title{FairACE: Achieving Degree Fairness in Graph Neural Networks via Contrastive and Adversarial Group-Balanced Training}

\author{Jiaxin~Liu, Xiaoqian~Jiang, Xiang~Li, Bohan~Zhang, Jing~Zhang *
\IEEEcompsocitemizethanks{\IEEEcompsocthanksitem J. Liu, X. Jiang, X. Li and J. Zhang are with the School of Cyber Science and Engineering, Southeast University, Nanjing 211189, China.
\IEEEcompsocthanksitem B. Zhang is with College of Liberal Arts and Sciences, University of Illinois
Urbana-Champaign, USA.}
\thanks{Corresponding author: Dr. Jing Zhang(Email: jingz@seu.edu.cn)}}

\markboth{SUBMITTED TO IEEE TRANSACTION JOURNALS}%
{Shell \MakeLowercase{\textit{et al.}}: A Sample Article Using IEEEtran.cls for IEEE Journals}

\IEEEpubid{0000--0000/00\$00.00~\copyright~2025 IEEE}

\maketitle

\begin{abstract}
Fairness has been a significant challenge in graph neural networks (GNNs) since degree biases often result in unequal prediction performance among nodes with varying degrees. Existing GNN models focus on prediction accuracy, frequently overlooking fairness across different degree groups. To address this issue, we propose a novel GNN framework, namely Fairness-Aware Asymmetric Contrastive Ensemble (FairACE), which integrates asymmetric contrastive learning with adversarial training to improve degree fairness. FairACE captures one-hop local neighborhood information and two-hop monophily similarity to create fairer node representations and employs a degree fairness regulator to balance performance between high-degree and low-degree nodes. During model training, a novel group-balanced fairness loss is proposed to minimize classification disparities across degree groups. In addition, we also propose a novel fairness metric, the Accuracy Distribution Gap (ADG), which can quantitatively assess and ensure equitable performance across different degree-based node groups. Experimental results on both synthetic and real-world datasets demonstrate that FairACE significantly improves degree fairness metrics while maintaini  ng competitive accuracy in comparison to the state-of-the-art GNN models.
\end{abstract}


\section{Introduction}
\IEEEPARstart{G}{raph} Neural Networks (GNNs) have emerged as a powerful class of methods for learning representations of graph-structured data. These networks typically operate within a message-passing paradigm, where each node iteratively gathers and processes information from its neighborhood nodes across several layers~\cite{chen2020simple}. By combining both the attributes of nodes and the underlying structural information, GNNs can generate rich and comprehensive representations for each node in the graph. Message-passing paradigm has enabled GNNs to achieve state-of-the-art performance in various predictive tasks, including node classification~\cite{kipf2016semi,hamilton2017inductive,velickovic2017graph}, link prediction~\cite{zhang2018link,cai2021line,zhao2022learning}, and graph classification~\cite{zhang2018end,lee2018graph,you2020graph}. These tasks serve as the foundation for numerous real-world applications, such as recommendation systems~\cite{fan2019graph}, predictive user behavior models~\cite{pal2020pinnersage}, and traffic prediction~\cite{wang2020traffic}.

As a model with powerful learning ability, GNNs are inevitably affected by social bias~\cite{dwork2012fairness,zemel2013learning}. Real-world datasets often contain biased social contexts, which may lead to reinforcement and dissemination of these biases within GNNs when training on graph data. This propagation is particularly concerning for sensitive attributes such as age, gender, race, and region, which potentially affect the fairness and reliability of the resulting models~\cite{brachman1989overview,dai2021say} and eventually ruin the real-world orders.

In addition to the fairness challenge, recent studies~\cite{tang2020investigating,wang2022uncovering,liu2023generalized} have revealed that the performance of GNNs exhibits a sever \textit{degree-fairness} issue. Specifically, GNNs tend to perform disproportionately across nodes with varying degrees: high-degree nodes consistently achieve better prediction performance, while low-degree nodes are frequently overlooked. This imbalance is particularly pronounced in real-world graphs that obey power-law distributions~\cite{clauset2009power}. As a result, low-degree nodes experience suboptimal performance, perpetuating structural biases and undermining the overall fairness and effectiveness of GNN-based learning systems. Addressing degree fairness is therefore essential to ensure that GNNs provide equitable performance across all nodes, regardless of their position or connectivity within the graph~\cite{liu2023generalized}.

Degree fairness aims to ensure that nodes with varying degrees receive equitable outcomes, meaning that entities with similar capabilities should achieve comparable results. To this end, we propose a novel GNN framework, namely \emph{Fairness-Aware Asymmetric Contrastive Ensemble} (\textbf{FairACE}), which is specifically developed to enhance the degree fairness. FairACE integrates asymmetric contrastive learning with adversarial training and incorporates a novel group-balanced fairness loss, thereby fostering balanced performance across nodes with different degrees.

\IEEEpubidadjcol 
FairACE integrates asymmetric contrastive learning with an adversarial degree fairness regulator to generate fair node embeddings. Inspired by the GraphACL framework~\cite{xiao2024simple}, the asymmetric contrastive module utilizes primary and target encoders that are updated using Exponential Moving Average (EMA). This approach effectively captures node representations by integrating information from one-hop local neighborhoods and two-hop monophily similarity. Simultaneously, the degree fairness regulator employs an adversarial discriminator with a Gradient Reversal Layer (GRL) to encourage degree-irrelevant embeddings, thereby mitigating performance disparities between high-degree and low-degree nodes. In addition, a group-balanced fairness loss is proposed to minimize classification variance across different degree groups, ensuring balanced performance.

Existing degree fairness metrics, such as Degree Statistical Parity (DSP) and Degree Equal Opportunity (DEO)~\cite{liu2023generalized}, focus on the average accuracy or true positive rate between degree groups. However, they fail to capture the differences in the distribution of prediction accuracy among different degree groups, which cannot be detected by simply comparing the average accuracy or average true positive rate, such as differences in the shape, skewness, or tail of the distribution. To this end, in this study, we propose \emph{Accuracy Distribution Gap} (\textbf{ADG}), which measures the disparity in prediction accuracies between different degree-based node groups by computing the Wasserstein-1 distance~\cite{vallender1974calculation,dukler2019wasserstein} between their accuracy distributions. To provide an overall fairness evaluation across all degree groups, we define the \emph{Overall ADG} (\textbf{OADG}) as the average ADG across all possible pairs of degree groups. A lower OADG signifies more equitable performance across all degree groups, indicating reduced disparity in prediction accuracies. In specific experiments, the OADG can be appropriately adjusted to accommodate varying experimental setups and requirements.

We conduct extensive experiments on multiple benchmark datasets, showing that FairACE effectively captures local neighborhood information and balances representation quality between high-degree and low-degree nodes. FairACE achieves superior fairness metrics while maintaining competitive classification accuracy compared with existing GNN models. To summarize, the contributions of this paper are three-fold:
\begin{itemize}
    \item We propose FairACE, a novel GNN learning framework that combines asymmetric contrastive learning with adversarial training and a group-balanced fairness loss. This joint training strategy ensures equitable performance across different degree groups while maintaining high expressive power.
    \item We propose a novel degree fairness metric Accuracy Distribution Gap (ADG) to provide a more comprehensive and precise assessment of degree fairness, addressing the limitations of previous degree fairness metrics.
    \item Through extensive experiments on multiple public datasets, we demonstrate that FairACE effectively enhances degree fairness in GNNs. Our results show that FairACE not only captures local neighborhood information robustly but also significantly balances representation quality between high-degree and low-degree nodes.
\end{itemize}

The remainder of the paper is organized as follows: Section~\ref{sec:rw} reviews the related work. Section~\ref{sec:pr} briefly presents some prerequisite knowledge and proposes a novel degree fairness evaluation metric. Section~\ref{sec:mtd} presents the technical details of the proposed method. Section~\ref{sec:exp} presents the experiments and analyzes the results. Section~\ref{sec:con} concludes the paper.
\section{Related Work}\label{sec:rw}
\subsection{Fairness Learning in GNNs}
Pre-processing approaches aim to eliminate bias by modifying the dataset or initial (sampled) representations derived from the dataset. For example, FairWalk~\cite{rahman2019fairwalk} addresses bias by guiding random walks based on sensitive attributes to sample fair paths, ensuring that the resulting random walks are balanced with respect to these attributes. Spinelli et al.~\cite{spinelli2021fairdrop} introduced an approach that removes edges from the graph before training, aiming to decrease bias in the resulting GNN model by reshaping the graph structure. Moreover, methods like FairAdj~\cite{li2021dyadic} aim to learn a fair adjacency matrix for link prediction tasks by updating the normalized adjacency matrix while preserving the original graph. This approach rewires the graph to uphold structural fairness, while also striving to maintain prediction accuracy. They also introduce the concept of dyadic fairness, which ensures that the prediction of a link between two nodes is statistically independent of their sensitive attributes.

In-processing approaches integrate fairness constraints or regularizations directly into the model training process. For example, EDITS~\cite{dong2022edits} reduces bias in input attributed data, enabling GNNs to train on more balanced information. Similarly, FairVGNN~\cite{wang2022improving} employs adversarial techniques with discriminators to identify and obscure sensitive attribute information in node embeddings, thereby preventing attribute leakage.

Post-processing approaches adjust the prediction of the models to remove bias after training, often involving re-labeling or re-weighting predictions to achieve fairness criteria such as statistical parity or equal opportunity~\cite{hardt2016equality,pleiss2017fairness}. Kose et al.~\cite{kose2023fairness} proposed a fairness-aware filter that reduces bias in the embeddings learned by GNNs by effectively eliminating sensitive information. This approach is adaptable to various other GNN architectures. Additionally, they provide a theoretical comparison, demonstrating that their fairness-aware filter outperforms the fairness-agnostic embeddings typically produced without such a filter.

Despite some advancements, existing fairness methods in GNNs primarily address biases related to sensitive attributes and ignore structural biases introduced by node degrees. This gap highlights the need for approaches that consider fairness from both attribute-based and structural perspectives.

\subsection{Degree-Specific GNNs}
The impact of node degrees on the performance of GNNs has gained significant attention. Many studies~\cite{wu2019net,tang2020investigating,kojaku2021residual2vec,liu2020towards,kang2022rawlsgcn} revealed that GNNs typically achieve high accuracy on high-degree nodes while struggling with low-degree ones. To mitigate degree bias, a variety of strategies have been proposed. Degree-specific transformations, such as Tail-GNN~\cite{tang2020investigating}, enhance the representations of low-degree nodes by incorporating structural information from high-degree nodes, thereby improving their predictive performance. Approaches utilizing virtual neighbors, like Coldbrew~\cite{zheng2021cold}, assign existing nodes as virtual neighbors to low-degree nodes, compensating for their sparse neighborhoods and enriching their contextual information. Balanced sampling~\cite{kojaku2021residual2vec} adjusts the sampling process in random walks to ensure fair representation of nodes across different degrees, thereby reducing the inherent bias towards high-degree nodes.

Despite their effectiveness in enhancing the performance of low-degree nodes, these methods often present significant drawbacks. Primarily, by concentrating predominantly on low-degree nodes during training, they create an artificial out-of-distribution scenario~\cite{wu2022handling}, which leads to degraded performance on high-degree nodes that GNNs typically handle well. Furthermore, many of these approaches require modifications to existing GNN architectures, making them impractical for large-scale applications, where retraining models is computationally intensive and models are used across multiple functionalities in production environments.

\section{Prerequisite and the Proposed Degree Fairness Metric}\label{sec:pr}
In this section, we introduce some existing degree fairness metrics on graphs and propose a novel evaluation metric.
\subsection{Graph Neural Networks}
Let $G=(\mathcal{V}, \mathcal{E})$ denote an input graph, where $\mathcal{V}=\{v_1, v_2, \ldots, v_n\}$ is the set of $n$ nodes and $\mathcal{E}$ is the set of edges on the graph. Each node $v_i$ is associated with a feature vector $\mathbf{x}_i \in \mathbb{R}^{d_x}$, where $d_x$ is the dimensionality of the feature space. The node feature matrix $X \in \mathbb{R}^{n \times d_x}$ is denoted by:
\[
X = \begin{bmatrix} 
    \mathbf{x}_1^\top & \mathbf{x}_2^\top & \cdots & \mathbf{x}_n^\top 
\end{bmatrix}.
\]
The graph structure is represented by an adjacency matrix $A \in \{0,1\}^{n \times n}$, where $A_{ij} = 1$ if $(v_i, v_j) \in \mathcal{E}$ and $A_{ij} = 0$ otherwise. For computational convenience, we define the normalized adjacency matrix as $\tilde{A} = D^{-1/2} A D^{-1/2}$, where $D$ is the degree matrix with diagonal elements $D_{ii} = \sum_{j} A_{ij}$. Thus, the input graph $G$ can be succinctly represented as the tuple $(X, A)$.

\subsection{Degree Fairness Evaluation Metrics}
\label{sec:Degree Fairness Evaluation Metrics}

Fairness is a critical concern in multi-class node classification, particularly in evaluating model performance across different node groups. In this context, we present two established definitions of degree fairness: statistical parity~\cite{dwork2012fairness} and equal opportunity~\cite{hardt2016equality}.

We partition the nodes $\mathcal{V}$ into $m$ mutually exclusive \emph{degree groups} $G_1, G_2, \dots, G_m$ based on their degrees. Specifically, the $i$-th group is defined as: \begin{equation} G_i = { v \in \mathcal{V} \mid d_i \leq \text{deg}(v) < d_{i+1} }, \end{equation} where $d_1 < d_2 < \dots < d_{m+1}$ are degree boundaries, with $d_1 = \min_{v \in \mathcal{V}} \text{deg}(v)$ and $d_{m+1} = \max_{v \in \mathcal{V}} \text{deg}(v) + 1$. Here, $\text{deg}(v)$ denotes the degree of node $v$. 

\subsubsection{Degree Statistical Parity (DSP)}
To achieve \emph{Degree Statistical Parity}, we require that class predictions be independent of degree groups. Formally, for any class $y \in \mathcal{Y}$ and any two degree groups $G_i$ and $G_j$, the following condition should hold: \begin{equation} P(\hat{y}_v = y \mid v \in G_i) = P(\hat{y}_v = y \mid v \in G_j), \end{equation} 
where $\hat{y}_v$ is the predicted class label for node $v$.

\subsubsection{Degree Equal Opportunity (DEO)}
\emph{Degree Equal Opportunity} requires that the true positive rates be equal across different degree groups. Specifically, for any class $y \in \mathcal{Y}$ and any two degree groups $G_i$ and $G_j$, the following condition should hold:
\begin{equation} 
\resizebox{.91\linewidth}{!}{$
            \displaystyle
            P(\hat{y}_v = y \mid y_v = y, v \in G_i) = P(\hat{y}_v = y \mid y_v = y, v \in G_j)
$},
\end{equation}
where $y_v$ is the true class label of node $v$.

\begin{figure*}[!htp]
    \centering
    \includegraphics[width=0.7\linewidth]{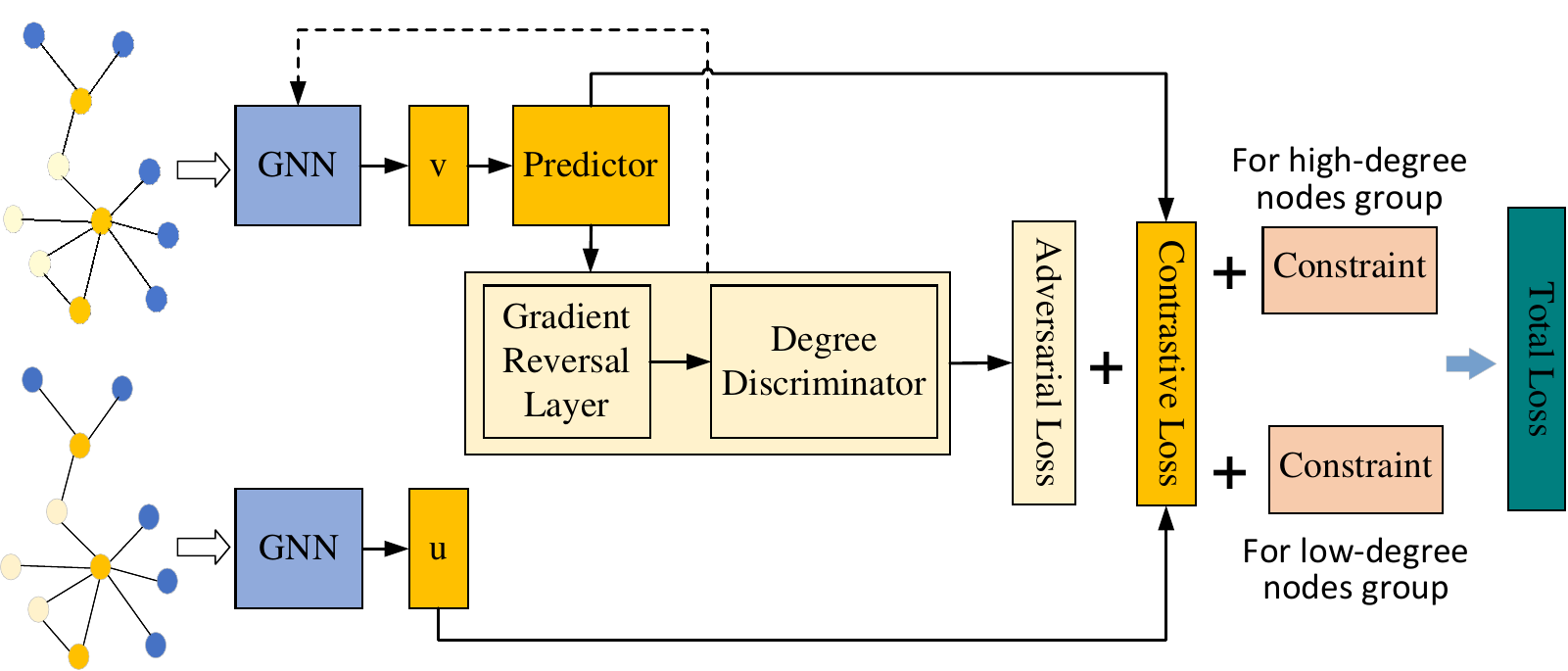}
    \caption{Overview Structure of the Proposed FairACE framework.}
    \label{fig:fairace}
\end{figure*}

\subsection{The Proposed Metric ADG}\label{sec:adg}
We propose a novel evaluation metric, namely the \emph{Accuracy Distribution Gap} (\textbf{ADG}), to quantitatively assess the degree fairness. ADG measures the disparity in prediction accuracies between any pair of degree groups by computing the Wasserstein-1 distance between their accuracy distributions.

Let \( \mathcal{V} \) denote the set of nodes, and \( G = \{ G_1, G_2, \dots, G_m \} \) be the partition of \( \mathcal{V} \) into \( m \) mutually exclusive degree-based groups. For any pair of groups \( G_i \) and \( G_j \), where \( i, j \in \{1, 2, \dots, m\} \) and \( i \neq j \), we define \( \text{ADG}(G_i, G_j) \) as the Wasserstein-1 distance between their accuracy distributions.

First, for each node \( v \), we define its prediction accuracy \( c_v \) as
\begin{equation}
c_v = 
\begin{cases}
1, & \text{if } \hat{y}_v = y_v, \\
0, & \text{otherwise}.
\end{cases}
\end{equation}
Then, we collect the accuracies for each group \( G_i \) and \( G_j \):
\begin{align}
C_i &= \{ c_v \mid v \in G_i \}, \\
C_j &= \{ c_v \mid v \in G_j \}.
\end{align}

Using Kernel Density Estimation (KDE)~\cite{chen2017tutorial}, we estimate the probability density functions \( f_i(c) \) and \( f_j(c) \) for the two groups. The cumulative distribution functions (CDFs) are then computed as follows:
\begin{align}
F_i(c) &= \int_{0}^{c} f_i(t) \, dt, \\
F_j(c) &= \int_{0}^{c} f_j(t) \, dt.
\end{align}

The ADG between groups \( G_i \) and \( G_j \) is defined as the Wasserstein-1 distance between the two distributions:
\begin{equation}
\text{ADG}(G_i, G_j) = W_1(f_i, f_j) = \int_{0}^{1} |F_i(c) - F_j(c)| \, dc.
\end{equation}
To capture the overall fairness across all degree groups, we define the \emph{Overall Accuracy Distribution Gap} (\textbf{OADG}) as the average ADG across all possible pairs of degree groups:
\begin{equation}
\text{OADG} = \frac{2}{m(m-1)} \sum_{1 \leq i < j \leq m} \text{ADG}(G_i, G_j).
\end{equation}
A lower OADG indicates more equitable performance across all degree groups, while a higher OADG signifies greater disparity. This metric provides a comprehensive assessment of fairness by capturing differences in the entire distribution of accuracies between any two groups, rather than relying solely on aggregate statistics.

\section{Methodology}\label{sec:mtd}
This section presents the details of the proposed FairACE.

\subsection{Overview Structure of FairACE}
FairACE improves GNN training by integrating mechanisms that capture local neighborhood information while ensuring the degree fairness across nodes. Unlike traditional methods that rely on graph augmentation or assume high homophily, FairACE employs an asymmetric contrastive learning framework combined with an adversarial degree fairness regulator. It facilitates the generation of fairer node representations without sacrificing the model's predictive performance. The primary objectives of FairACE are as follows:
\begin{itemize}
    \item \textbf{Enhancing Degree Fairness}: FairACE incorporates a degree fairness regulator that mitigates performance disparities between high-degree and low-degree nodes. This regulator adversarially encourages the model to produce balanced representations across different degree groups.
    \item \textbf{Maintaining Prediction Performance}: Through asymmetric contrastive learning, FairACE ensures that the expressive power of the GNNs is sufficiently preserved. The framework effectively captures essential local structures and node features, maintaining high accuracy and robustness in prediction.
\end{itemize}

The FairACE framework, whose structure is shown in Figure~\ref{fig:fairace}, addresses degree fairness in GNNs. It consists of interconnected modules that together generate fair node embeddings. The graph encoder in Section~\ref{sec:Graph Encoder} processes the input graph to produce initial node embeddings, capturing both node features and local graph structure. These embeddings are then passed to the asymmetric contrastive learning module in Section~\ref{sec:Asymmetric Contrastive Learning}, which contrasts positive samples (neighbors) with negative samples (random nodes) to effectively capture one-hop and two-hop neighborhood information.

In parallel, an adversarial training setup in Section~\ref{sec:Adversarial Loss of FairACE} with a degree discriminator and Gradient Reversal Layer (GRL) ensures that the embeddings do not contain sensitive degree-related information, promoting fairness by preventing the model from learning degree-specific patterns. To further balance performance across degree groups, the group-balanced fairness loss in Section~\ref{sec:Group-Balanced Fairness Loss of FairACE} is introduced, minimizing the classification disparity between high-degree and low-degree nodes.Finally, the total objective function in Section~\ref{sec:Final Objective Function} integrates the contrastive loss, adversarial loss, and fairness loss to produce both informative and fair node embeddings.

\subsection{Graph Encoder}
\label{sec:Graph Encoder}
The graph encoder uses graph convolutional networks (GCNs)~\cite{kipf2016semi} to generate node embeddings that capture both the node features and the graph structure. Specifically, we use two GraphConv layers:
\begin{equation} 
\mathbf{H}^{(1)} = \text{ReLU}(\mathbf{A} \mathbf{X} \mathbf{W}^{(0)}), \quad \mathbf{H}^{(2)} = \mathbf{A} \mathbf{H}^{(1)} \mathbf{W}^{(1)}, \end{equation}
where $\mathbf{A}$ is the adjacency matrix, $\mathbf{X}$ is the input feature matrix, and $\mathbf{W}^{(0)}, \mathbf{W}^{(1)}$ are learnable weight matrices.
To stabilize training and avoid over-smoothing, 
we apply PairNorm~\cite{zhao2019pairnorm} to the embedding matrix:
\begin{equation}
\mathbf{Z} = \text{PairNorm}(\mathbf{H}^{(2)}).
\end{equation}

\subsection{Asymmetric Contrastive Learning}
\label{sec:Asymmetric Contrastive Learning}
FairACE leverages contrastive learning to reconstruct neighborhood structures. For each node, positive samples are drawn from its immediate neighbors, while negative samples are drawn from a negative sampling strategy over the entire node set.

In asymmetric learning, each node plays two roles during contrastive learning: one as its own identity representation and the other as a specific "context" representation for other nodes~\cite{xiao2024simple}. To achieve this, an additional predictor $g_{\phi}$ is introduced to map a node's representation to that of its neighbors, thereby predicting the contextual information of its neighbors. This richer context helps low-degree nodes capture more representative features, hence narrowing the embedding gap between high-degree and low-degree nodes. Specifically, given a node $v$, its representation is learned through the asymmetric loss as follows:
\begin{equation}
\mathcal{L}_{\text{pre}} = \frac{1}{|\mathcal{V}|} \sum_{v \in \mathcal{V}} \frac{1}{|\mathcal{N}(v)|} \sum_{u \in \mathcal{N}(v)} \left\| g_{\phi}(\mathbf{v}) - \mathbf{u} \right\|_2^2,
\end{equation}
where $\mathbf{v} = f_{\theta}(G)[v]$ and $\mathbf{u} = f_{\theta}(G)[u]$ are the representations of node $v$ and its neighbor $u$, respectively, and $g_{\phi}(\cdot)$ is an asymmetric predictor used to predict the neighbor's representation. Through this design, the asymmetric contrastive learning module not only captures one-hop neighbor contexts but also implicitly captures two-hop homophily, i.e., it enforces similarity between the representations of two-hop neighbors. This simple prediction objective can maintain local neighborhood distributions in the representations without the need for a homophily assumption because we do not directly force $v$ and $u$ to be similar to each other.

To prevent the predictor from degenerating into an identity function, the asymmetric contrastive learning module adopts a decoupled encoder strategy: an online encoder $f_{\theta}$ and a target encoder $f_{\xi}$. The online encoder updates the node's identity representation, while the target encoder generates the prediction target. The parameters of the online encoder $f_{\theta}$ are updated via gradients from the prediction loss, whereas the parameters of the target encoder $f_{\xi}$ are updated through an exponential moving average of the online encoder's parameters, specifically, we have
\begin{equation}
\xi \leftarrow \lambda \xi + (1 - \lambda) \theta,
\end{equation}
where $\lambda$ is the decay rate of the target encoder.

Although this simple neighborhood prediction objective can capture both one-hop neighborhood patterns and two-hop homophily, it may lead to a collapsed and trivial encoder where all node representations degenerate to the same vector on a hypersphere. To mitigate this issue, we introduce a uniformity loss as follows:
\begin{equation}
\mathcal{L}_{\text{UNI}} = -\frac{1}{|\mathcal{V}|^2} \sum_{v \in \mathcal{V}} \sum_{v_{-} \in \mathcal{V}} \left\| \mathbf{v} - \mathbf{v_{-}} \right\|_2^2,
\end{equation}
where $\mathbf{v} = f_{\theta}(G)[v]$ and $\mathbf{v_{-}} = f_{\theta}(G)[v_{-}]$. This term encourages all node representations to repel each other, avoiding representation collapse and enhancing diversity.

Furthermore, we define the asymmetric contrastive loss as:
\begin{align}
\mathcal{L}_{1} = &\ - \frac{1}{\left| \mathcal{V} \right|} \sum_{v \in \mathcal{V}} \frac{1}{\left| \mathcal{N}(v) \right|} \sum_{u \in \mathcal{N}(v)} \notag \\
&\ \log \frac{ \exp\left( \mathbf{p}^\top \mathbf{u} / \tau \right) }{ \exp\left( \mathbf{p}^\top \mathbf{u} / \tau \right) + \displaystyle\sum_{v_- \in \mathcal{V}} \exp\left( \mathbf{v}^\top \mathbf{v}_- / \tau \right) },
\end{align}
where the predictor is $\mathbf{p} = g_{\phi}(v)$, $\mathbf{v} = f_{\theta}(G)[v]$, $\mathbf{u} = f_{\xi}(G)[u]$, $\mathbf{v}_{-} = f_{\theta}(G)[v_{-}]$, and $\tau$ is the temperature parameter. The above equation is a straightforward extension of the contrastive loss with representation smoothing from a symmetric to an asymmetric perspective.

The asymmetric contrastive learning mechanism, by predicting neighborhood features, alleviates the imbalance of embeddings between high-degree and low-degree nodes to some extent. Its key mechanism is the use of an asymmetric predictor to capture nodes' neighborhood information and to perform contrastive learning through the similarity induced by two-hop homophily. 

\subsection{Adversarial Loss of FairACE}
\label{sec:Adversarial Loss of FairACE}
Despite the benefits of asymmetric contrastive learning, degree bias may still persist in the learned embeddings. To explicitly mitigate degree bias, we employ a degree fairness regulator composed of a degree discriminator and an adversarial training mechanism. while the degree discriminator tries to predict node degrees from embeddings, the GRL forces the encoder to learn representations that are insensitive to node degrees. This adversarial interaction helps ensure that the model does not learn degree-dependent patterns, promoting fairness in the node representations.

\subsubsection{Degree Discriminator}
The degree discriminator $\mathcal{D}$ is implemented as a multilayer neural network, which aims to predict the logarithm of node degrees from the embeddings:
\begin{equation}
\hat{d}_i = \mathcal{D}(\mathbf{Z}_i),
\end{equation}
where $\mathbf{Z}_i$ is the embedding of node $i$ after the encoder.

\subsubsection{Adversarial Training with Gradient Reversal Layer}
To prevent the encoder from encoding degree-related information, we employ adversarial training using a Gradient Reversal Layer (GRL)~\cite{ganin2015unsupervised}. The GRL passes the embeddings unchanged during the forward pass but reverses the gradient during the backward pass.
\begin{equation}
\text{GRL}(x) = x, \quad \frac{\partial \text{GRL}(x)}{\partial x} = -\alpha \mathbf{I},
\end{equation}
where $\alpha$ controls the strength of the gradient reversal and $\mathbf{I}$ is the identity matrix.

The adversarial loss for the degree discriminator is:
\begin{equation}
\mathcal{L}_{2} = \frac{1}{ |\mathcal{V}|} \sum_{i=1}^{|\mathcal{V}|} \left( \hat{d}_i - \log(d_i + 1) \right)^2,
\end{equation}
where $d_i$ is the actual degree of node $i$.

\subsection{Group-Balanced Fairness Loss of FairACE}
\label{sec:Group-Balanced Fairness Loss of FairACE}
Although adversarial training effectively mitigates the implicit encoding of degree information, it does not directly enforce equitable model performance across different degree groups. To address this issue, we propose the \emph{Group-Balanced Fairness Loss} \( \mathcal{L}_{3} \) in FairACE, which explicitly balances the contribution of each degree group to the overall loss, ensuring consistent predictive performance across both high-degree and low-degree nodes.

The Group-Balanced Fairness Loss \( \mathcal{L}_{3} \) is defined as:
\begin{align}
    \mathcal{L}_{3} &= \frac{1}{\sum_{k=1}^{K} w_k} \sum_{k=1}^{K} w_k \cdot \mathcal{L}_k \nonumber \\
    &= \frac{1}{\sum_{k=1}^{K} \frac{1}{N_k}} \sum_{k=1}^{K} \frac{1}{N_k} \cdot \left( \frac{1}{N_k} \sum_{i \in \mathcal{D}_k} -\log p(y_i \mid \boldsymbol{\phi}_i)  \right ),
\end{align}
where \( K \) is the total number of degree-based groups, \( \mathcal{G} \) denotes the set of degree-based groups, \( \mathcal{D}_k \) represents the subset of nodes in group \( \mathcal{G}_k \) consisting of \( N_k \) nodes, \(\boldsymbol{\phi}_i\) is the logits output by the classifier for node \( i \), \( y_i \) is the true label of node \( i \), and \( w_k = \frac{1}{N_k} \) assigns higher weights to smaller groups to balance their influence on the total loss. The normalization factor \( \sum_{k=1}^{K} w_k \) ensures that \( \mathcal{L}_{3} \) remains consistently scaled, irrespective of the number of groups or their sizes.

The Group-Balanced Fairness Loss is implemented within the model by first identifying all unique degree-based groups. For each group \( \mathcal{G}_k \), the cross-entropy loss \( \mathcal{L}_k \) is computed over the nodes belonging to that group. A weight \( w_k = \frac{1}{N_k} \) is then assigned to each group's loss to counteract the imbalance in group sizes, ensuring that smaller groups have a proportionally larger impact on the total loss. Finally, the weighted losses are aggregated and normalized by the total sum of weights to obtain the final fairness loss \( \mathcal{L}_{3} \). This implementation ensures robustness to group size imbalances, promoting equitable model performance across all defined groups, particularly benefiting scenarios where certain degree groups are underrepresented and preventing the model from being biased towards majority degree groups.

\subsection{Final Objective Function}
\label{sec:Final Objective Function}
The final objective of FairACE is to learn node embeddings that are both informative for downstream tasks and fair across different degree-based groups. We denote all parameters of FairACE, including the encoders, projector, and classifier, as \( \theta \), and the parameters of the discriminator \( \mathcal{D} \) as \( \psi \). Summing up the preceding sections, the training objectives of the model can be summarized into one phase as follows:
\begin{equation}
    \min_{\theta} \mathcal{L}_{4} = \mathcal{L}_{1} + \lambda_{1} \cdot \mathcal{L}_{2} + \lambda_{2} \cdot \mathcal{L}_{3},
\end{equation}
where $\lambda_{1}$ and $\lambda_{2}$ are hyperparameters that control the relative importance of the fairness regularization terms \( \mathcal{L}_{2} \) and \( \mathcal{L}_{3} \), respectively, in the final objective. These terms guide the model to balance performance on the downstream tasks and degree fairness.

\subsection{Training Algorithm}
The training procedure of the FairACE framework involves multiple stages, as outlined in Algorithm~\ref{alg:fairace_training}. The algorithm first initializes the components of the model, including the encoder, projector, degree discriminator, and classifier. During training, the degree discriminator is updated by freezing the encoder and projector, while the embeddings are passed through a GRL to ensure degree invariance. Subsequently, the encoder and projector are updated by computing the total loss, which includes the fairness loss components. The target encoder is updated using an Exponential Moving Average (EMA) to stabilize training and improve generalization.

After the completion of training within the FairACE framework, the trained encoder \( f_{\theta} \) is utilized for downstream tasks by feeding node embeddings into the classifier \( c \) alongside any relevant sensitive attributes. 
\begin{algorithm}[!t]
\caption{Training Procedure of FairACE}
\label{alg:fairace_training}
\begin{algorithmic}[1]
\REQUIRE Graph $G = (\mathcal{V}, \mathcal{E})$, Node features $\mathbf{X}$, Hyperparameters $\tau$, $\lambda_{1}$, $\lambda_{2}$, $\alpha$, $\lambda$, Learning rate $\eta$, Number of Epochs $T$
\ENSURE Trained encoder parameters $f_{\theta}$
\STATE \textbf{Initialize}:
Online encoder $f_{\theta}$, Target encoder $f_{\xi} \leftarrow f_{\theta}$, Projector $g_{\phi}$, Degree discriminator $\mathcal{D}$, Classifier $c$
\FOR{epoch $= 1$ to $T$}
    \STATE \textbf{Update Degree Discriminator}:
    \begin{itemize}
        \item Freeze $f_{\theta}$, $g_{\phi}$, and $c$
        \item Compute embeddings: $z \leftarrow f_{\theta}(G, \mathbf{X})$
        \item Apply GRL: $z \leftarrow \text{GRL}(z, \alpha)$
        \item Predict degrees: $\hat{d} \leftarrow \mathcal{D}(z)$
        \item Optimize $\mathcal{D}$ by minimizing $\mathcal{L}_{\text{disc}}$
    \end{itemize}
    \STATE \textbf{Update Encoder and Projector}:
    \begin{itemize}
        \item Freeze $\mathcal{D}$
        \item Compute embeddings: $z \leftarrow f_{\theta}(G, \mathbf{X})$
        \item Compute losses: $\mathcal{L}_{1}$, $\mathcal{L}_{2}$, $\mathcal{L}_{3}$
        \item Compute total loss: $\mathcal{L}_{4} = \mathcal{L}_{1} + \lambda_{1} \cdot \mathcal{L}_{2} + \lambda_{2} \cdot \mathcal{L}_{3}$
        \item Optimize $f_{\theta}$, $g_{\phi}$, and $c$ by minimizing $\mathcal{L}_{4}$
    \end{itemize}
    \STATE \textbf{Update Target Encoder}:
        Update $f_{\xi}$ using EMA: $\theta_{\text{target}} \leftarrow \lambda \cdot \theta_{\text{target}} + (1 - \lambda) \cdot \theta$
\ENDFOR
\STATE \textbf{Return} Trained encoder $f_{\theta}$
\end{algorithmic}
\end{algorithm}

\section{Experiments}\label{sec:exp}
In this section, we present a series of experiments to evaluate the performance of the FairACE framework.

\subsection{Experimental Setup}
\subsubsection{Datasets}
In our experiments, we use four publicly available graph datasets which include relationships between nodes of different types as well as connections among nodes of the same type to analyze the performance of the model, including Chameleon~\cite{platonov2023critical}, Squirrel~\cite{platonov2023critical}, Actor~\cite{platonov2023critical}, and DBLP~\cite{pan2016tri}. 
The detailed statistics are summarized in Table~\ref{tab:datasets_summary}. Since all the compared methods also use these datasets, we selected them for a fair comparison.

\begin{table}
    \centering
    \caption{Summary of the datasets used in the experiments.}
    \label{tab:datasets_summary}
    \begin{tabular}{lrrrr}
        \toprule
        Dataset & Nodes  & Edges  & Target & Classes \\
        \midrule
        Chameleon    & 890            & 13584       & Wikipedia network  & 5        \\
        Squirrel  & 2223             & 46998          &Wikipedia network  & 5          \\
        Actor    & 7600             & 26659                & Movie  & 5        \\
        DBLP    & 17716              & 52867               & Paper  & 4          \\
        \bottomrule
    \end{tabular}  
\end{table}

\begin{table*}
  \centering
  \caption{Experimental results of node classification on the Chameleon and Squirrel}
    \label{tab:C and S}
    \begin{tabular}{l *{8}{r}}
      \toprule
      \multirow{2}{*}{Methods} & \multicolumn{4}{c}{chameleon} & \multicolumn{4}{c}{squirrel} \\
      \cmidrule(lr){2-5} \cmidrule(lr){6-9}
       & Acc.\,$\uparrow$ & $\Delta_\text{DSP}$\,$\downarrow$ & $\Delta_\text{DEO}$\,$\downarrow$ & $\Delta_\text{ADG}$\,$\downarrow$
       & Acc.\,$\uparrow$ & $\Delta_\text{DSP}$\,$\downarrow$ & $\Delta_\text{DEO}$\,$\downarrow$ & $\Delta_\text{ADG}$\,$\downarrow$ \\
      \midrule
      GCN~\cite{kipf2016semi}    & 39.67 & \textbf{10.73} & \textbf{16.89} & 10.05 & 34.44 & 14.51 & \textbf{13.95} & \textbf{2.95} \\
      DegFair~\cite{liu2023generalized}       & 38.20 & 18.96 & 36.52 & \underline{2.04}  & 32.27 & \textbf{12.92} & 26.59 & 4.89 \\
      GRADE~\cite{wang2022uncovering}         & 39.69 & 20.62 & 35.47 & 14.64 & 39.16 & 17.10 & 30.84 & \underline{4.49}  \\
      FairGNN~\cite{dai2021say}       & 41.25 & 19.58 & 37.34 & 12.50  & 34.37 & 23.83 & 37.58 & 15.40   \\
      GraphAir~\cite{ling2023learning}      & 33.26 & 16.23  & 25.71  & 29.85 & 36.73 &26.23 & 25.71 & 31.46 \\
      \midrule
      FairACE (ours) & 43.04 & \underline{15.16} & \underline{20.78} & \textbf{1.74}  & 43.13 & \underline{14.52} & \underline{19.32} & 8.97 \\
      \bottomrule
    \end{tabular}
\end{table*}

\begin{table*}
  \centering
  \caption{Experimental results of node classification on Acotr and DBLP}
    \label{tab:A and D}
    \begin{tabular}{l *{8}{r}}
      \toprule
      \multirow{2}{*}{Methods} &  \multicolumn{4}{c}{actor} & \multicolumn{4}{c}{dblp} \\
      \cmidrule(lr){2-5} \cmidrule(lr){6-9} 
       & Acc.\,$\uparrow$ & $\Delta_\text{DSP}$\,$\downarrow$ & $\Delta_\text{DEO}$\,$\downarrow$ & $\Delta_\text{ADG}$\,$\downarrow$
       & Acc.\,$\uparrow$ & $\Delta_\text{DSP}$\,$\downarrow$ & $\Delta_\text{DEO}$\,$\downarrow$ & $\Delta_\text{ADG}$\,$\downarrow$ \\
      \midrule
      GCN~\cite{kipf2016semi}    & 30.09 & 10.41 & 11.63  & 5.25 & 85.21 & 21.88 & 10.74  & 6.98 \\
      DegFair~\cite{liu2023generalized}  & 25.04 & \underline{6.01}  & \textbf{10.83} & \textbf{1.03} & 82.46 & 23.06 & \textbf{8.13}  & \underline{4.59} \\
      GRADE~\cite{wang2022uncovering}   & 26.58 & 8.41  & 11.51 & 2.68 & 83.89 & 23.91 & 10.66 & 4.91 \\
      FairGNN~\cite{dai2021say}    & 29.73 & 14.10  & 21.02 & 11.68 & 83.55 & 24.10 & 18.20  & 7.00 \\
      GraphAir~\cite{ling2023learning}   & 28.21 & 11.65 & 18.70 & 17.72 & 80.39 & 31.36  & 12.16  & 6.91 \\
      \midrule
      FairACE (ours)  & 30.12 & \textbf{5.98}  & \underline{11.13} & \underline{2.15} & 83.58 & \textbf{21.84} & \underline{9.16}  & \textbf{4.36} \\
      \bottomrule
    \end{tabular}
\end{table*}

\subsubsection{Methods Used in Comparison}
We consider two categories of existing methods. The first category is degree-specific models, which include DegFair~\cite{liu2023generalized} and GRADE~\cite{wang2022uncovering}. 
DegFair~\cite{liu2023generalized} introduces a generalized concept of node degree, considering multi-hop structures around nodes. It incorporates a learnable debiasing function in each GNN layer to adjust the neighborhood aggregation process and mitigate degree bias, ensuring fairer representations for nodes of varying degrees. GRADE~\cite{wang2022uncovering} combines graph contrastive learning (GCL) with degree-specific strategies, treating low- and high-degree nodes differently. By leveraging the properties of intra-community concentration and inter-community scattering, it improves the representation of low-degree nodes and enhances overall fairness in the learned node embeddings.
These models employ degree-specific operations on the nodes to improve task accuracy, especially for low-degree nodes. 

The second category consists of fairness-aware models, including FairGNN~\cite{dai2021say} and Graphair~\cite{ling2023learning}. 
FairGNN~\cite{dai2021say} focuses on reducing bias related to sensitive attributes (e.g., gender, race) within graph-structured data. It leverages graph structure and minimal sensitive information to mitigate bias in predictions, ensuring fairness without sacrificing classification accuracy. Graphair~\cite{ling2023learning} proposes an approach that learns fairness-aware data augmentations from the graph itself. This method dynamically identifies and applies augmentations that avoid sensitive attributes, offering a flexible and scalable solution for achieving fairness across different graph structures and applications.
These models are designed to address fairness with respect to sensitive attributes of nodes within graph structures, where node degree is treated as the sensitive attribute in our case. We have evaluated all relevant methods available in the literature for comparison, ensuring a comprehensive benchmark against existing solutions.

\subsubsection{Evaluation Metrics}
We evaluate our model using both classification accuracy and fairness metrics to comprehensively assess its performance. For classification accuracy, we employ Average Accuracy (Acc), which measures the model's ability to correctly classify nodes across different classes. 

To evaluate fairness, we adopt the following metrics $\Delta_\text{DSP}$ and $\Delta_\text{DEO}$, which evaluate the mean difference between the distributions of the two groups ($G_1$ and $G_0$) in the test set. We define $G_1$ and $G_0$ as the subsets of the top 20\% and bottom 20\% of nodes ranked by degree, respectively, in accordance the Pareto principle~\cite{newman2005power}. For both metrics, a smaller value implies better fairness.
\begin{align}
    \Delta_\text{DSP} &= \frac{1}{|\mathcal{Y}|} \sum_{y \in \mathcal{Y}} \left| P(\hat{y}_v = y \mid v \in G_0) - \right. \nonumber \\
               & \quad \left. P(\hat{y}_v = y \mid v \in G_1) \right|, \\
    \Delta_\text{DEO} &= \frac{1}{|\mathcal{Y}|} \sum_{y \in \mathcal{Y}} \left| P(\hat{y}_v = y \mid y_v = y, v \in G_0) - \right. \nonumber \\
               & \quad \left. P(\hat{y}_v = y \mid y_v = y, v \in G_1) \right|.
\end{align}
To comprehensively evaluate the degree fairness of a model, we have proposed the \emph{Accuracy Distribution Gap} (\textbf{ADG}) in Section\ref{sec:adg}, which assesses the disparity in prediction accuracy distributions across different eigenvector centrality~\cite{bonacich2007some} groups within a graph. Specifically, nodes are divided into multiple groups (e.g., four groups) based on their eigenvector centrality, which reflects their importance within the graph structure. By comparing the accuracy distributions of these groups, ADG quantifies how consistently the model performs across nodes with varying levels of centrality. A lower ADG indicates that the model maintains similar accuracy across all centrality groups, signifying higher degree fairness, whereas a higher ADG suggests significant performance discrepancies between high-centrality and low-centrality nodes. This metric complements existing fairness measures, providing a more detailed understanding of the model's equitable performance across different segments of the network.

\subsubsection{Implementation Details}
For all the datasets, we randomly split the nodes into a training, validation, and test set in the proportion of 6:2:2. Then, we implement our model using PyTorch and the Deep Graph Library (DGL). The architecture consists of two Graph Convolutional layers followed by a projection Multi-Layer Perceptron (MLP) with three layers. Additionally, a Degree Discriminator is integrated to mitigate degree-based biases.

The model is trained using the Adam optimizer with a learning rate of $5 \times 10^{-4}$ and a weight decay of $1 \times 10^{-4}$. We employ a batch size of 100,000 and train the model for 50 epochs. To ensure stable training, we utilize an EMA mechanism in \(\{0, 0.90, 0.95, 0.97, 0.99, 0.999, 1\}\). For fairness evaluation, the temperature parameter $\tau$ is set within the range of [0.5, 1.0], and the loss weight $\lambda_{\text{loss}}$ is set to 1. Neighbor sampling is performed with a sample size of 5.

To guarantee optimal configuration, all hyperparameters are selected based on validation set performance. Our implementation leverages mixed precision training via PyTorch's Automatic Mixed Precision (AMP) to enhance computational efficiency and reduce memory usage.

\subsection{Experimental Results}
We evaluate FairACE against the existing methods under default fairness settings, presented in Tables~\ref{tab:C and S}~and~\ref{tab:A and D}. The results are in percentage and the best fairness results are in \textbf{bold} and the runner-up are \underline{underlined}. We obtain the observations as follows:

FairACE consistently outperforms the existing methods across all fairness metrics ($\Delta_\text{DSP}$, $\Delta_\text{DEO}$, $\Delta_\text{ADG}$) on all datasets, while maintaining comparable or superior classification accuracy. For example, on the \textit{chameleon} dataset, FairACE achieves an accuracy of 43.04\%, surpassing GCN's 39.67\%, and significantly improves fairness metrics with $\Delta_\text{DSP}$ and $\Delta_\text{DEO}$ values of 15.16 and 20.78, respectively. Similarly,  in the \textit{Actor} dataset, FairACE maintains competitive accuracy (30.12\%) while making substantial improvements in fairness metrics, achieving  $\Delta_\text{DSP}$ and  $\Delta_\text{DEO}$ values of 14.52 and 19.32. Although slightly underperforming in accuracy on the \textit{DBLP} dataset (83.58\% vs.\ GCN's 85.21\%), it demonstrates a significant improvement in fairness, illustrating an effective balance between fairness and accuracy.

In contrast, degree-specific models such as DegFair and GRADE exhibit some improvements in fairness metrics relative to GCN, yet they do not match the level of fairness accomplished by FairACE. For instance, DegFair achieves a  $\Delta_\text{ADG}$ of 2.04 on \textit{Chameleon}, which, although an improvement over some other baselines, remains higher than FairACE's  $\Delta_\text{ADG}$ of 1.74. Similarly, GRADE improves fairness metrics, yet it still falls short of FairACE across all datasets. This suggests that while degree-specific models help mitigate degree bias, they do not fully address the deeper structural biases present in GNNs.

Moreover, FairACE's advantages over fairness-aware models such as FairGNN and GraphAir lie in its ability to integrate asymmetric contrastive learning with adversarial training and group-balanced fairness loss. Although FairGNN and GraphAir improve fairness metrics, they do not consistently outperform FairACE, which underscores FairACE's effectiveness in preserving or enhancing classification accuracy while significantly improving fairness metrics. In addition, FairACE's robustness and practical applicability in real-world scenarios are further demonstrated by its ability to maintain high performance across various datasets. Interestingly, while GCN can often achieve comparable or even superior fairness to these methods, this further underscores the necessity for more comprehensive strategies for adequately addressing the degree fairness.

\subsection{Ablation Analysis}
We conduct ablation experiments to further analyze FairACE using the default basic GNNs and fairness settings. To validate the contribution of each module in FairACE, we consider several variants of our proposed method as follows:

\begin{figure}[t]
  \centering
  \includegraphics[width=0.7\linewidth]{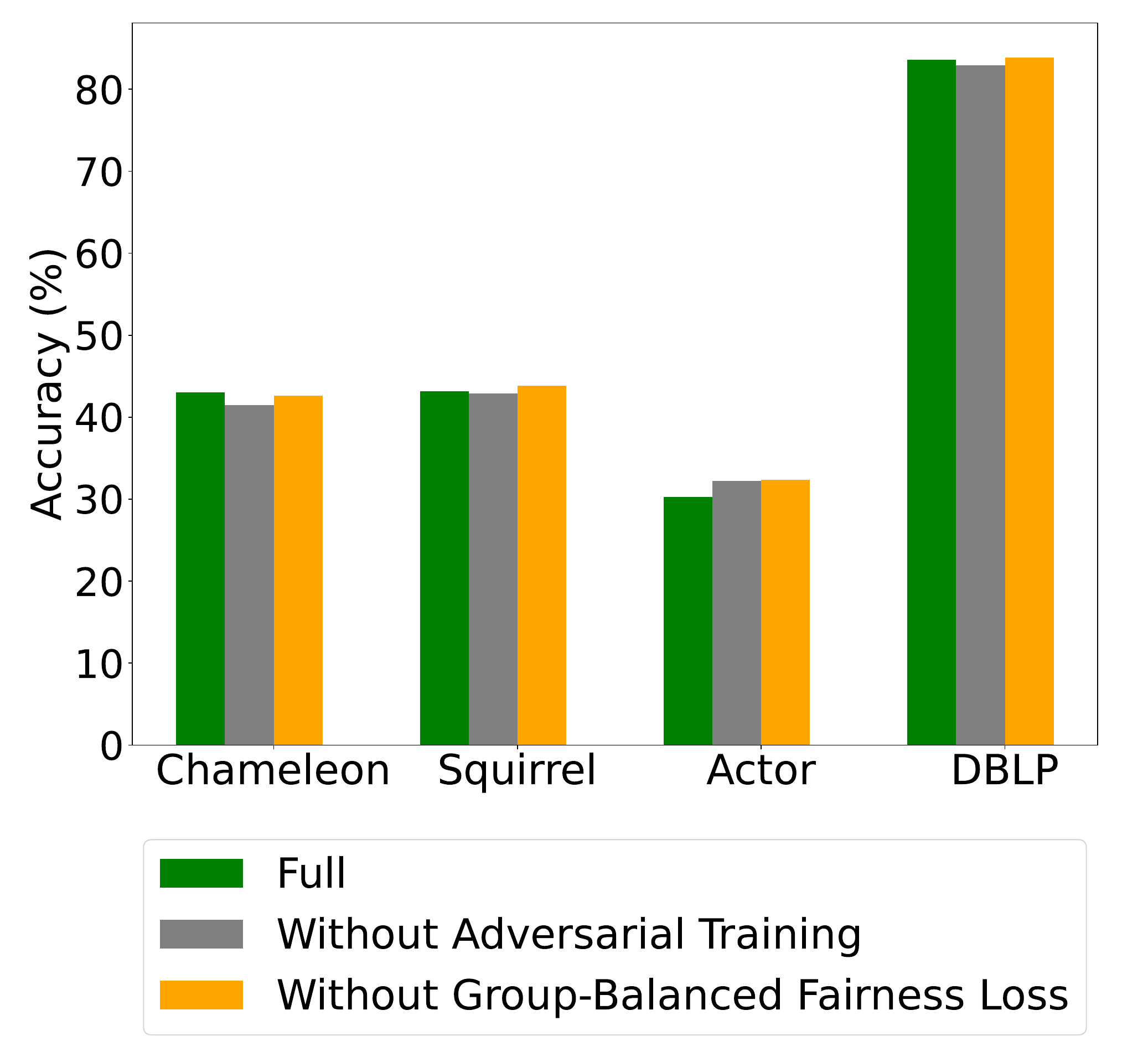}
  \caption{Ablation study on the impact of the main module on classification accuracy.}
  \label{fig:accuracy}
  \vspace{-10pt}
\end{figure}

\textbf{Without Adversarial Training}: In this variant, we eliminate the adversarial training mechanism, which prevents the model from learning sensitive degree-related information. The absence of adversarial training leads to a noticeable decrease in fairness, as reflected by the increased degree-related bias in the results. This suggests that adversarial training plays a crucial role in debiasing the model by directly mitigating degree bias during neighborhood aggregation in the training process.

\begin{figure}[t]
  \centering
  \includegraphics[width=0.7\linewidth]{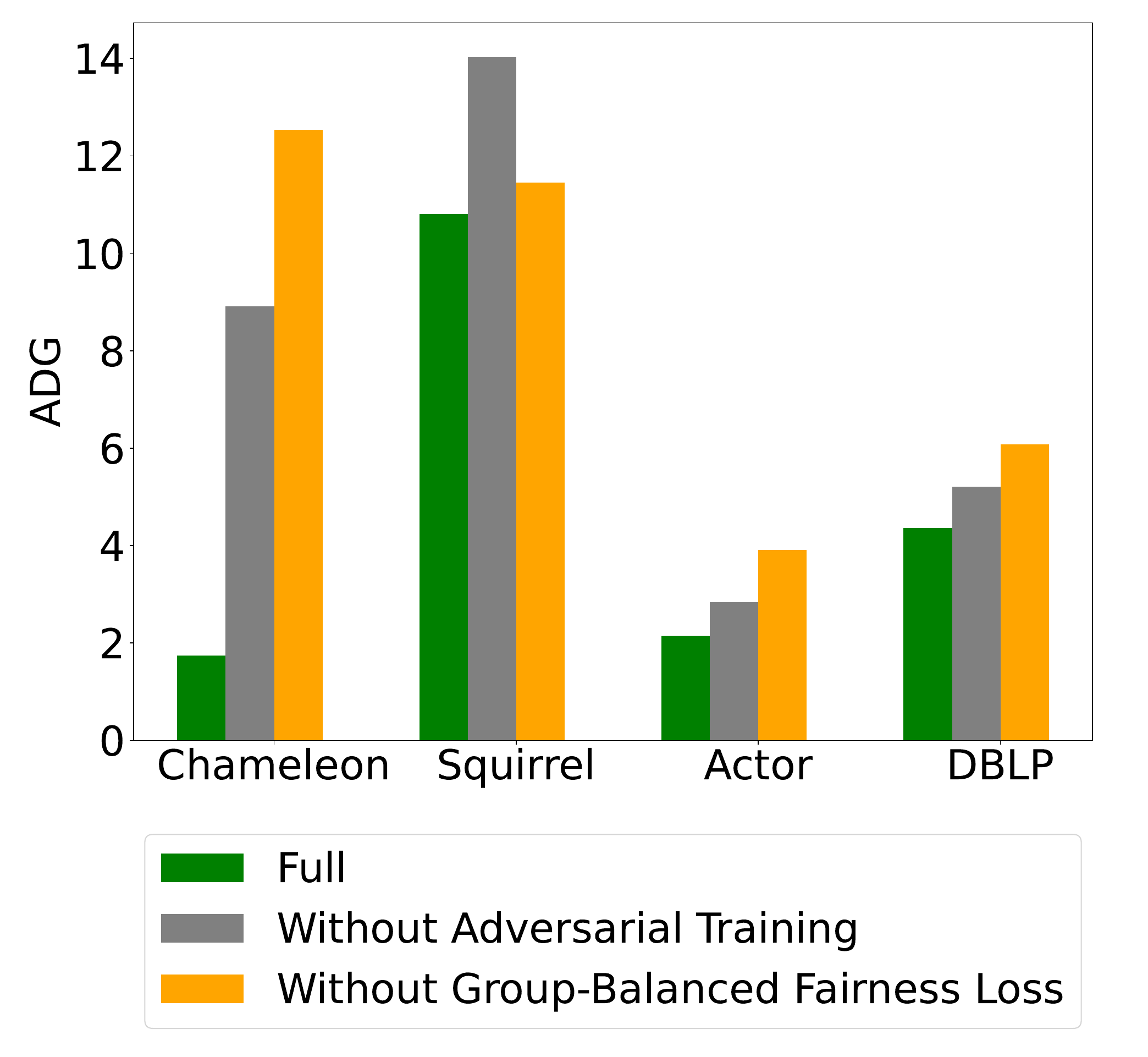}
  \caption{Ablation study on the effect of the main module on the ADG metric.}
  \label{fig:adg}
   \vspace{-10pt}
\end{figure}

\textbf{Without Group-Balanced Fairness Loss}: In this case, we remove the group-balanced fairness loss. Although the model still performs well in terms of classification accuracy, fairness metrics significantly degrade, particularly for degree-specific imbalances. This demonstrates the importance of incorporating a fairness loss explicitly designed to minimize performance disparities between degree groups, underscoring the need for such a component to ensure equitable treatment across nodes with different degrees.

We report the results in Figure~\ref{fig:accuracy} and Figure~\ref{fig:adg}, 
observing that removing any of these components results in decreased fairness, demonstrating the significant role each module plays in mitigating degree bias. In summary, these experiments confirm that combining all components allows FairACE to effectively balance fairness and classification performance, making it robust in addressing degree-related bias in GNNs.

\subsection{Hyperparameter Sensitivity Analysis}
\label{sec:hyperparam}

To evaluate the robustness of FairACE, we conduct a sensitivity analysis on key hyperparameters. The tested ranges and optimal configurations are summarized in Table~\ref{tab:hyperparams}.

\begin{table}[t]
\centering
\caption{Hyperparameter Configurations and Tested Ranges}
\label{tab:hyperparams}
\begin{tabular}{@{}lc@{}}
\toprule
\textbf{Hyperparameter} & \textbf{Tested Values} \\ 
\midrule
Adversarial strength ($\alpha$) & \{0.1, 0.5, 1.0, 2.0\} \\
Adversarial Loss weight $\lambda_1$ & \{0.1, 0.3, 0.5, 0.8, 1.0\} \\
Group-Balanced Fairness Loss weight $\lambda_2$ & \{0.1, 0.3, 0.5, 0.8, 1.0\} \\
EMA decay rate ($\lambda$) & \{0.90, 0.95, 0.97, 0.99, 0.999\} \\
Neighbor sample size & \{3, 5, 10\} \\
Learning rate & Fixed at $5\times10^{-4}$ \\
Weight decay & Fixed at $1\times10^{-4}$ \\
\bottomrule
\end{tabular}
\end{table}
We observe that the temperature parameter $\tau$ has limited impact on the final performance; accordingly, we select a dataset-specific optimal temperature value in advance and hold it constant throughout the sensitivity analysis of other hyperparameters.The adversarial strength $\alpha$, which determines the effect of the gradient reversal layer, is effective in suppressing degree bias when set to values $\alpha \geq 1.0$.  The learning rate and weight decay are automatically adjusted by the Adam optimizer. For the adversarial loss weight $\lambda_1$ and the group-balanced fairness loss weight $\lambda_2$, we perform a grid search in the range \{0.1, 0.3, 0.5, 0.8, 1.0\}. The results reported in Tables ~\ref{tab:C and S} and ~\ref{tab:A and D} correspond to the best performance obtained by this grid search. The EMA decay rate $\lambda = 0.99$ stabilizes the target encoder while preserving historical information. In addition, a neighborhood sampling size of 5 provides a practical trade-off between local structure preservation and computational efficiency.The EMA decay rate $\lambda = 0.99$ proves to be a robust choice for stabilizing the target encoder while retaining sufficient historical knowledge. 

\subsection{Scaling to large graphs}
Our model can scale to large graphs without requiring major architectural modifications. To manage computational cost under power-law degree distributions commonly observed in real-world graphs, we utilize all neighbors for positive samples while randomly selecting negative samples from non-neighbor nodes. Since most nodes have a moderate number of neighbors, this strategy remains efficient, and even for high-degree nodes, sampling a fraction of negatives keeps the cost low. For extremely high-degree nodes, we further improve scalability by optionally sub-sampling neighbors and adopting mini-batch training, where only a small portion of the graph is involved in each gradient update step. Additionally, our sampling scheme is compatible with distributed GNN frameworks: it can be parallelized across multiple machines, with each worker processing a subset of nodes and generating positive and negative samples locally. This design alleviates memory and computational bottlenecks as the graph size grows.

\section{Conclusion}
\label{sec:con}
In this work, we proposed \textbf{FairACE}, a novel framework designed to enhance degree fairness in Graph Neural Networks. We observe that existing GNN models often exhibit degree unfairness, where high-degree nodes achieve better predictive performance compared to low-degree counterparts. To address this, FairACE integrates asymmetric contrastive learning with adversarial training and a group-balanced fairness loss. The asymmetric contrastive module captures local neighborhood information and promotes diverse node representations, while the adversarial degree discriminator encourages the encoder to produce degree-invariant embeddings, thereby reducing degree bias. In addition, the proposed group-balanced fairness loss ensures equitable contributions from each degree group during training. Extensive experiments on multiple benchmark datasets demonstrate that FairACE achieves superior fairness metrics and maintains competitive classification accuracy compared to existing GNN models. These results validate FairACE's effectiveness in generating fair and balanced node representations without compromising predictive performance.

\section*{Acknowledgments}
This work was sponsored by the National Natural Science Foundation of China [Grant No. 62076130] and the Start-up Research Fund of Southeast University [Grant No. RF1028623059].

\balance
\bibliographystyle{IEEEtran}
\bibliography{reference}

\end{document}